\pdfoutput=1
\documentclass[11pt]{article}
\usepackage[]{acl}
\usepackage[T1]{fontenc}
\usepackage[utf8]{inputenc}
\usepackage{times}
\usepackage{latexsym}
\usepackage{microtype}
\usepackage{inconsolata}
\usepackage{graphicx}
\usepackage{booktabs}
\usepackage{multirow}
\usepackage{siunitx}
\usepackage{amsmath}
\usepackage{array}
\usepackage{enumitem}
\usepackage{makecell}
\usepackage{tikz}
\usepackage{graphicx}
\usetikzlibrary{positioning,arrows.meta,calc,fit,shapes.geometric,shapes.multipart}
\usepackage{cleveref}

\title{LLMs on Tabular Data with Limited Semantics: \\Evidence from Industrial Car Retrofit Prediction}
\author{ Aina Vila Pons$^{1,2}$ \quad Ioannis Tzachristas$^{1}$ \quad Constantinos Antoniou$^{1}$ \\ $^{1}$Chair of Transportation Systems Engineering, Technical University of Munich, Germany \\ $^{2}$BMW Group, Munich, Germany }

\newcommand{\LLMEmbed}{\textsc{LLM-Embed}}
\newcommand{\LLMPrompted}{\textsc{LLM-Prompted}}
\newcommand{\LLMStacked}{\textsc{LLM-Stacked}}

\begin{document}
\maketitle

\begingroup
\renewcommand\thefootnote{}
\footnotetext{This research was conducted as part of the industrial Master's Thesis of Aina Vila Pons at the Technical University of Munich, in collaboration with BMW Group, under the co-supervision of her BMW manager, Mr. Guido Klöss.}
\endgroup

\begin{abstract}
Industrial retrofit planning depends on structured operational data rather than free text: planners must estimate whether a newly registered prototype will require a retrofit, which retrofit package it will need, and how long the work will take.
We study an industrial dataset linking a prototype-registration system (284{,}271 vehicles) with a retrofit-management system (48{,}716 cleaned visits), and compare strong tabular machine learning baselines with three LLM-based strategies on row-serialized inputs: embedding features (Amazon Titan), direct prompted classification (Claude Sonnet~4), and an ML+LLM stacking approach.
Across binary occurrence prediction, 15-way retrofit-type classification, per-visit duration regression, and an aggregated monthly benchmark, classical tree ensembles remain the strongest standalone models.
However, the LLM results reveal a consistent pattern: embeddings remain useful on tables (binary AUC $=0.982$), direct prompting collapses once semantic signal is stripped by hashing (binary AUC $=0.500$; multiclass weighted F1 $=0.018$), and hybrid stacking yields the best manually built multiclass model (weighted F1 $=0.626$).
On the monthly benchmark, lag-based machine learning outperforms time-series foundation models, though Chronos-small remains competitive in zero-shot forecasting.
The results suggest that on privacy-constrained industrial tables, LLMs are more effective as complementary components than as replacements for strong tabular baselines.
\end{abstract}

\section{Introduction}
Retrofit departments in automotive development modify prototype vehicles by implementing hardware and software changes so that downstream testing can proceed.
Capacity planning is difficult because early vehicle metadata is high-dimensional, mostly categorical, and noisy, while the retrofit demand signal is rare.
In practice, planners need answers to three questions: \emph{will} a newly registered prototype visit the retrofit department, \emph{what} retrofit package will it need, and \emph{how long} will the retrofit likely take?

We study that pipeline through the lens of LLMs over structured data.
The inputs are tables whose categorical values are hashed, so row serialization exposes structure but very little lexical semantics.
That makes the setting useful for separating what LLM-based methods can learn from distributional patterns alone from what they need semantic content to solve.
We compare strong tabular baselines against three LLM-based strategies on serialized rows, and we also benchmark time-series foundation models on an aggregated monthly planning signal. \footnote{\url{https://github.com/aina-vila-pons/retrofit-forecast-pipeline}.}

\paragraph{Contributions.}
\begin{itemize}[leftmargin=*,itemsep=1pt,topsep=2pt]
    \item We formulate an end-to-end industrial planning problem over structured data: binary occurrence prediction, 15-way retrofit-type prediction, per-visit duration regression, and a monthly time-series benchmark.
    \item We compare three LLM-based strategies for tabular prediction---embedding features, prompted classification, and ML+LLM stacking---against strong classical baselines and an AutoML reference.
    \item We report a clear failure mode: direct prompting performs poorly, while embedding-based features and hybrid stacking still retain useful signal.
    \item We distill deployment lessons on cost, latency, and privacy-constrained model design for LLM systems operating over enterprise tables.
\end{itemize}

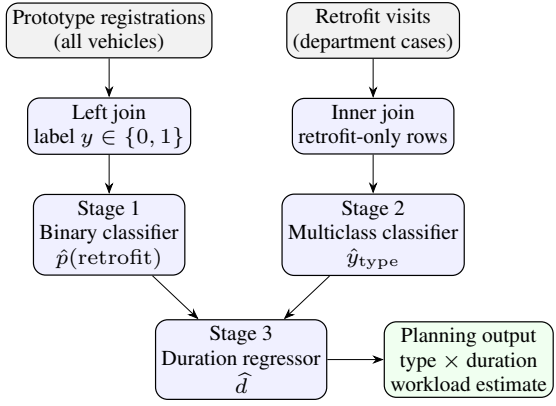
\begin{figure}[t]
  \centering
  \resizebox{0.94\columnwidth}{!}{%
  \begin{tikzpicture}[
    font=\scriptsize,
    box/.style={draw, rounded corners, align=center,
                minimum height=0.55cm, minimum width=1.55cm, inner sep=2pt},
    data/.style={box, fill=gray!10},
    stage/.style={box, fill=blue!6},
    outbox/.style={box, fill=green!6},
    arr/.style={-{Stealth[length=1.3mm]}, line width=0.3pt}
  ]

  \node[data] (reg)
    {Prototype registrations\\(all vehicles)};
  \node[data, right=0.85cm of reg] (retro)
    {Retrofit visits\\(department cases)};

  \node[stage, below=0.45cm of reg] (leftjoin)
    {Left join\\label $y\in\{0,1\}$};
  \draw[arr] (reg) -- (leftjoin);

  \node[stage, below=0.45cm of retro] (innerjoin)
    {Inner join\\retrofit-only rows};
  \draw[arr] (retro) -- (innerjoin);

  \node[stage, below=0.45cm of leftjoin] (stage1)
    {Stage~1\\Binary classifier\\$\hat p(\mathrm{retrofit})$};
  \draw[arr] (leftjoin) -- (stage1);

  \node[stage, below=0.45cm of innerjoin] (stage2)
    {Stage~2\\Multiclass classifier\\$\hat y_{\mathrm{type}}$};
  \draw[arr] (innerjoin) -- (stage2);

  \coordinate (mid12) at ($(stage1.south)!0.5!(stage2.south)$);
  \node[stage, below=0.5cm of mid12] (stage3)
    {Stage~3\\Duration regressor\\$\widehat{d}$};
  \draw[arr] (stage1) -- (stage3);
  \draw[arr] (stage2) -- (stage3);

  \node[outbox, right=0.65cm of stage3] (out)
    {Planning output\\type $\times$ duration\\workload estimate};
  \draw[arr] (stage3) -- (out);

  \end{tikzpicture}%
  }
  \caption{Compact view of the three-stage planning pipeline. LLM methods are evaluated at the classification stages by serializing each structured row into text.}
  \label{fig:pipeline}
\end{figure}

\section{Related work}
\paragraph{LLMs for structured data.}
Recent work studies whether language-model interfaces can improve prediction on tables.
One direction treats a row serialization as text and learns from its embedding representation \cite{dinh2022lift,hegselmann2023tabllm}.
A second direction performs direct prompted prediction over serialized rows, optionally with few-shot examples \cite{hegselmann2023tabllm}.
A third direction combines LLM outputs with classical tabular models in hybrid systems.
In parallel, tabular foundation models such as TabPFN ask whether pretrained transformers can match boosted trees on structured data \cite{hollmann2023tabpfn}.

\paragraph{Industrial tabular ML and temporal forecasting.}
For high-cardinality industrial tables, target encoding \cite{micci2001preprocessing} and gradient-boosted trees such as XGBoost, LightGBM, and CatBoost remain strong baselines \cite{chen2016xgboost,ke2017lightgbm,prokhorenkova2018catboost}.
Rare-event prediction further requires imbalance-aware training and careful metric selection \cite{saito2015precisionrecall,chawla2002smote}.
For the temporal side, we compare classical statistical forecasting \cite{box1970timeseries,hyndman2021fpp3,taylor2018prophet} with recent time-series foundation models including Chronos and TIME-LLM \cite{ansari2024chronos,jin2024timellm}.

\section{Data and problem formulation}
\subsection{Data sources}
We use two internal systems from a large automotive manufacturer:
(1) a \emph{prototype-registration system} that records all newly built prototype vehicles with early metadata such as derivative, integration step, build phase, engine type, build location, and deadlines; and
(2) a \emph{retrofit-management system} that records retrofit visits with retrofit package labels and process timestamps.
All categorical values are hashed; continuous features used in this work are engineered from timestamps, fleet counts, and frequency statistics.
For LLM-based methods, each row is additionally serialized into a key--value description that can be embedded or passed to a prompted classifier.

\subsection{Joins and labels}
We derive two supervised datasets.
A left join produces the binary occurrence dataset ($n{=}284{,}271$), where the target indicates whether the vehicle ever visits the retrofit department (positive rate approximately $4.4\%$; 12{,}378 positives vs.\ 271{,}893 negatives).
An inner join produces the retrofit-only dataset ($n{=}54{,}174$ visits before duration filtering), used for retrofit-type prediction and duration regression.
After grouping 17 rare multi-type combinations into their rarest component, the retrofit-type task contains 15 classes.
For duration, we remove P1--P99 outliers (503 rows), leaving 52{,}507 visits.
A compact dataset overview is given in Appendix~\ref{app:dataset}.

\paragraph{Targets.}
\textbf{Stage~1: occurrence.} Given a registration row $x$, predict whether the vehicle will come for retrofit.
\textbf{Stage~2: retrofit type.} For retrofit visits, predict the retrofit-type label.
\textbf{Stage~3: duration.} Predict the number of days between retrofit start and end date.
The raw duration distribution is right-skewed (mean $=9.2$ days, median $=4.0$ days), so we apply a $\log(1{+}x)$ transform for modeling.

\paragraph{Aggregated monthly benchmark.}
In addition to the per-visit duration task, we construct a 76-point monthly time series of mean retrofit duration.
This benchmark provides a coarse planning signal and a deliberately low-data setting for time-series foundation models.
It also lets us compare vehicle-level supervision against a much smaller aggregated forecasting problem.

\section{Methods}
\begin{figure*}[t]
  \centering
  \includegraphics[width=0.98\textwidth]{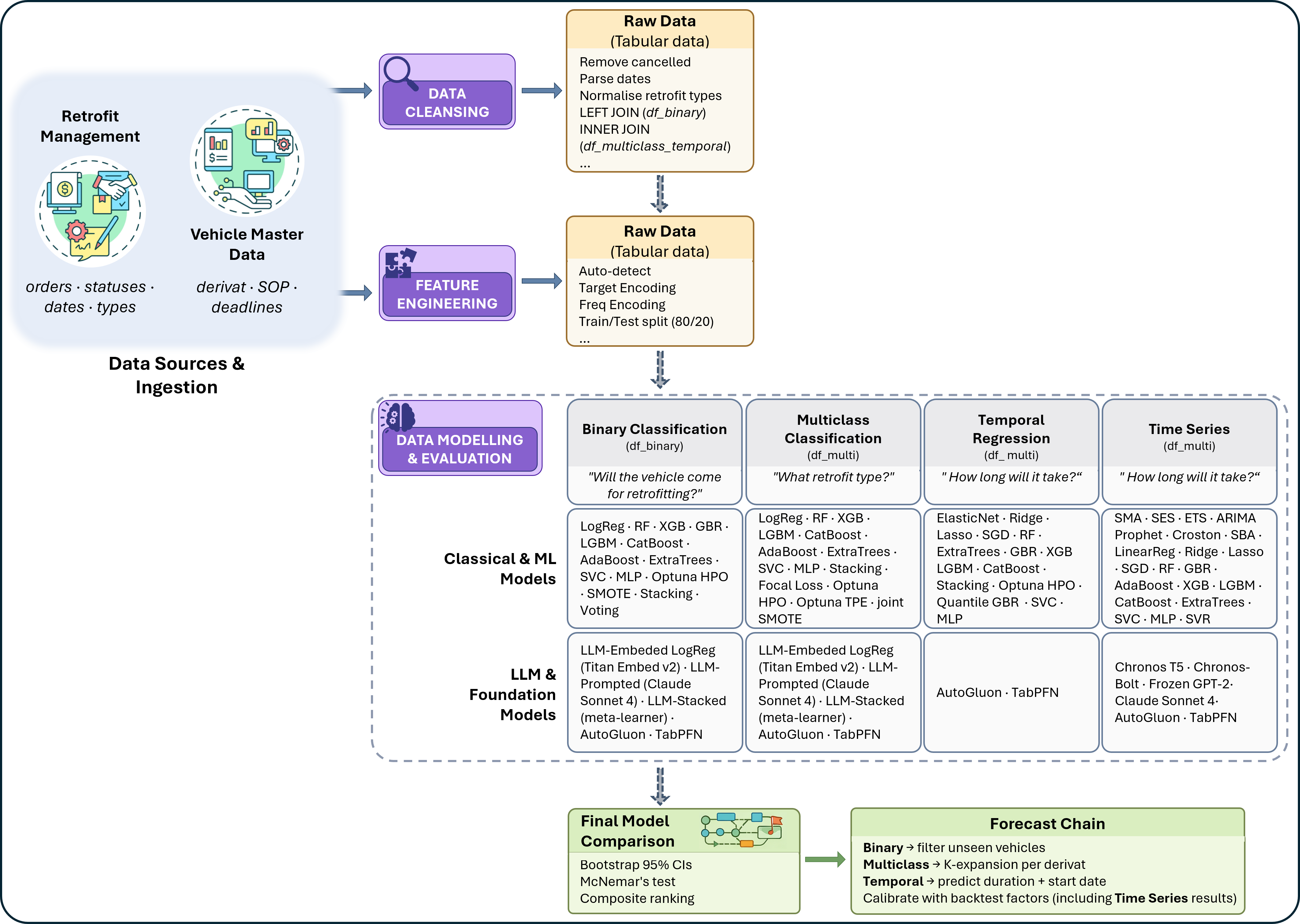}
  \caption{Complete modeling pipeline: from data ingestion through classical machine learning, LLM-based methods, and time-series foundation-model benchmarks to final comparison and forecast chaining.}
  \label{fig:pipeline-overview}
\end{figure*}

Figure~\ref{fig:pipeline-overview} summarizes the complete modeling pipeline.
Starting from the two operational databases, the workflow proceeds through data cleansing, feature engineering, and three supervised learning stages.
Each classification stage is evaluated with both classical machine learning models and LLM-based alternatives, while AutoML and time-series foundation models provide additional reference points.

\subsection{Feature engineering and classical baselines}
We distinguish categorical metadata from engineered numeric signals.
Numeric features include vehicle age in months, fleet-size counts aggregated by derivative, base type, engine type, and build location, as well as interaction terms such as age$\times$fleet and priority$\times$fleet.
To mitigate leakage, we exclude direct identifiers and temporal fields that would trivially reveal the target.
After target encoding, frequency encoding, and importance-based pre-filtering, the binary task retains 44 features and the multiclass/temporal tasks retain 50 features each.

Across the three stages we evaluate logistic regression, Random Forest \cite{breiman2001randomforests}, Extra Trees, Gradient Boosting, XGBoost, LightGBM, and CatBoost \cite{chen2016xgboost,ke2017lightgbm,prokhorenkova2018catboost}.
For Stage~1 we also train stacking and weighted soft-voting ensembles from the top-performing models.
Hyperparameters are tuned with Optuna \cite{akiba2019optuna}; for the multiclass task, the SMOTE ratio is tuned jointly with model hyperparameters \cite{chawla2002smote}.

\subsection{LLM-based methods}
We evaluate three LLM-based strategies at the classification stages.

\paragraph{\LLMEmbed.}
Each structured row is serialized into a key--value text string and embedded with Amazon Titan Embed~v2 (1{,}024 dimensions).
A logistic regression classifier is trained on the resulting vectors.
Because of API cost, the embedding stage is subsampled to 5{,}000 training rows.

\paragraph{\LLMPrompted.}
Claude Sonnet~4 receives each serialized row together with task-specific few-shot examples and predicts the label directly.
We use direct prompted classification rather than fine-tuning.
Because of cost and latency, prompted inference is evaluated on smaller held-out subsets (200--500 rows depending on task).

\paragraph{\LLMStacked.}
A meta-learner combines the best classical probabilities with LLM-derived outputs.
This design tests whether LLMs are more useful as feature providers or complementary components than as end-to-end predictors.

\subsection{AutoML and time-series baselines}
AutoGluon serves as an AutoML reference across the tabular tasks \cite{erickson2020autogluon}.
For the monthly benchmark, we compare statistical baselines, machine learning on lag features, AutoGluon on lag features, and time-series foundation models including Chronos and TIME-LLM \cite{ansari2024chronos,jin2024timellm}.
The benchmark is intentionally small and sparse, making it a useful stress test for zero-shot forecasting.

\section{Experimental setup}
We use an 80/20 stratified train--test split for classification and regression, and a 16-month holdout for the monthly time series.
We additionally compute bootstrap confidence intervals and pairwise McNemar tests; the main text reports the core performance tables, while detailed confidence intervals for the binary task are retained in Appendix~\ref{app:fulltables}.

Evaluation metrics are ROC-AUC, PR-AUC, and thresholded F1 for Stage~1; accuracy, weighted F1, macro F1, and one-vs-rest AUC for Stage~2; and MAE, RMSE, and $R^2$ for Stage~3.
Because Stage~1 is heavily imbalanced, ROC-AUC is interpreted jointly with PR-AUC \cite{saito2015precisionrecall}.
For the monthly benchmark we report MAE and $R^2$ on the 16-month test window.

\section{Results}
\subsection{Stage~1: occurrence prediction}
\Cref{tab:binary-main} shows the main binary results.
Hyperparameter-tuned CatBoost achieves the strongest standalone performance (AUC $=0.997$, F1 $=0.884$), while the voting ensemble yields the best PR-AUC.
Among the LLM-based methods, embeddings remain useful on rows (AUC $=0.982$), but direct prompting collapses to random performance (AUC $=0.500$).
The hybrid stack attains F1 $=0.900$ on its 200-row evaluation subset, suggesting complementary signal, but the overall picture remains clear: strong tabular models dominate as standalone predictors.

\begin{table}[t]
\centering
\small
\setlength{\tabcolsep}{3.8pt}
\caption{Stage~1 binary occurrence prediction. $\dagger$ indicates the smaller 200-row LLM evaluation subset. Full table in Appendix~\ref{app:fulltables}.}
\label{tab:binary-main}
\begin{tabular}{lccc}
\toprule
Model & ROC-AUC & PR-AUC & F1 \\
\midrule
CatBoost (HP-tuned) & \textbf{0.997} & 0.932 & 0.884 \\
Voting ensemble & 0.997 & \textbf{0.937} & 0.883 \\
\LLMStacked{}$^\dagger$ & 0.996 & 0.897 & \textbf{0.900} \\
\LLMEmbed{} & 0.982 & 0.667 & 0.684 \\
\LLMPrompted{}$^\dagger$ & 0.500 & 0.045 & 0.086 \\
AutoGluon & 0.997 & --- & 0.881 \\
\bottomrule
\end{tabular}
\end{table}

\subsection{Stage~2: retrofit-type prediction}
The multiclass task provides the clearest positive result for hybrid LLM use.
As shown in \Cref{tab:multi-main}, \LLMStacked{} reaches the best weighted F1 among the manually configured pipelines (0.626), slightly ahead of Random Forest with SMOTE (0.621) and XGBoost (0.614).
AutoGluon remains best overall at 0.654 weighted F1.
Again, direct prompted classification performs poorly on the inputs, while embedding-based features remain serviceable but clearly below the top tabular baselines.

\begin{table}[t]
\centering
\small
\setlength{\tabcolsep}{3.8pt}
\caption{Stage~2 retrofit-type prediction. Full table in Appendix~\ref{app:fulltables}.}
\label{tab:multi-main}
\begin{tabular}{lccc}
\toprule
Model & Acc & F1$_w$ & F1$_m$ \\
\midrule
AutoGluon & \textbf{0.702} & \textbf{0.654} & 0.348 \\
\LLMStacked{} & 0.680 & 0.626 & 0.281 \\
Random Forest (SMOTE) & 0.611 & 0.621 & \textbf{0.390} \\
XGBoost (SMOTE) & 0.601 & 0.614 & 0.380 \\
\LLMEmbed{} & 0.480 & 0.521 & 0.327 \\
\LLMPrompted{} & 0.100 & 0.018 & 0.012 \\
\bottomrule
\end{tabular}
\end{table}

\subsection{Stage~3: duration prediction and monthly benchmark}
For per-visit duration, classical ensembles still dominate: AutoGluon achieves MAE $=4.9$ days and Extra Trees achieves MAE $=5.1$ days.
On the monthly series, lag-based machine learning remains best (LightGBM and AutoGluon both reach MAE $=3.16$), but Chronos-small is reasonably competitive at MAE $=4.03$ without task-specific training.
TIME-LLM performs worse than Chronos but remains in the range of simple statistical baselines.
These results suggest that for the temporal component, foundation models are promising complements but not yet replacements for compact supervised baselines.

\begin{table}[t]
\centering
\small
\setlength{\tabcolsep}{4.0pt}
\caption{Stage~3 duration results. The full monthly benchmark is in Appendix~\ref{app:timeseriesdetails}.}
\label{tab:time-main}
\begin{tabular}{llcc}
\toprule
Setting & Model & MAE & $R^2$ \\
\midrule
\multirow{2}{*}{Per-visit} & AutoGluon & \textbf{4.9} & \textbf{0.38} \\
& Extra Trees & 5.1 & 0.37 \\
\midrule
\multirow{4}{*}{Monthly} & LightGBM (lags) & \textbf{3.16} & -1.90 \\
& AutoGluon (lags) & 3.16 & -1.88 \\
& Chronos-small & 4.03 & -0.07 \\
& TIME-LLM & 4.62 & -0.22 \\
\bottomrule
\end{tabular}
\end{table}

\section{Discussion and deployment considerations}
\paragraph{Operating point selection.}
Stage~1 models achieve ROC-AUC values of at least 0.996, but operational utility depends on the chosen threshold.
In practice, false positives waste effort by triggering unnecessary preparation, while false negatives create capacity risk.
We therefore recommend selecting operating points using precision--recall trade-offs rather than treating ROC-AUC as sufficient on its own.

\paragraph{LLM integration trade-offs.}
The LLM results show a clear boundary.
Once categorical values are hashed, direct prompted classification loses the semantic cues that make language-model reasoning useful.
By contrast, row embeddings still preserve co-occurrence and frequency structure, and the stacking model can exploit complementary error patterns.
For practitioners, that means LLMs are currently more promising as components inside a broader tabular system than as end-to-end predictors on enterprise tables.

The LLM components also impose real deployment costs.
Embedding 5{,}000 rows required minutes of API time, and prompted classification on a few hundred rows was slower still.
When target encoding and boosted trees already perform strongly, those costs need a clear return.
In this dataset, the strongest justification for LLM integration is complementary signal rather than raw leaderboard improvement.

\paragraph{Temporal forecasting and drift.}
Chronos-small and TIME-LLM provide usable forecasts without task-specific training, but the best lag-based models remain stronger on this short series.
Prototype programs also change quickly: new derivatives, build phases, and retrofit types appear regularly.
That makes drift monitoring and periodic retraining necessary regardless of whether the downstream model is classical or foundation-based.

\section{Conclusion}
We presented a multi-stage industrial planning study over enterprise tables.
Across tasks, the main pattern is consistent: on privacy-constrained structured data, direct prompted classification is brittle, embedding features remain useful, and hybrid ML+LLM systems can improve selected tasks without displacing strong tabular baselines.
Taken together, the results suggest that current LLM methods are most effective as complementary components for structured industrial prediction.

\bibliography{refs}

\appendix
\clearpage
\section{Additional dataset overview}
\label{app:dataset}
\begin{table}[h]
\centering
\small
\setlength{\tabcolsep}{4pt}
\caption{Dataset overview and derived learning tasks.}
\resizebox{\columnwidth}{!}{%
\begin{tabular}{@{}lrr@{}}
\toprule
Dataset / task & Rows & Vehicles \\
\midrule
Registration system (raw) & 284{,}271 & 284{,}271 \\
Retrofit system (cleaned) & 48{,}716 & 11{,}830 \\
Stage~1: binary (left join) & 284{,}271 & 284{,}271 \\
Stage~2: multiclass (inner join) & 54{,}174 & --- \\
Stage~3: duration (inner join subset) & 52{,}507 & --- \\
\bottomrule
\end{tabular}%
}
\end{table}

\section{Full classification tables}
\label{app:fulltables}
Tables~\ref{app:binaryfull} and~\ref{app:multifull} provide the complete binary and multiclass comparisons, including the stronger manual baselines omitted from the main text for space.

\begin{table}[h]
\centering
\small
\setlength{\tabcolsep}{3pt}
\caption{Full binary occurrence results (Stage~1). F1 is reported at the optimal threshold.}
\label{app:binaryfull}
\resizebox{\columnwidth}{!}{%
\begin{tabular}{lcccc}
\toprule
Model / approach & ROC-AUC & PR-AUC & F1@opt & AUC CI \\
\midrule
CatBoost (HP-tuned) & \textbf{0.997} & 0.932 & 0.884 & [0.997, 0.997] \\
Voting ensemble (top-3) & 0.997 & \textbf{0.937} & 0.883 & [0.997, 0.998] \\
Extra Trees (SMOTE) & 0.997 & 0.935 & 0.880 & [0.997, 0.998] \\
XGBoost (HP-tuned) & 0.996 & 0.913 & 0.884 & [0.996, 0.997] \\
LightGBM (HP-tuned) & 0.996 & 0.910 & 0.883 & [0.996, 0.997] \\
\LLMStacked{} & 0.996 & 0.897 & 0.900 & [0.987, 1.000] \\
Random Forest (HP-tuned) & 0.997 & 0.929 & 0.879 & [0.996, 0.997] \\
LogReg (feat-selected) & 0.990 & 0.810 & 0.777 & [0.989, 0.991] \\
\LLMEmbed{} LogReg & 0.982 & 0.667 & 0.684 & [0.977, 0.986] \\
AutoGluon (Ens.\ L3) & 0.997 & --- & 0.881 & --- \\
\LLMPrompted{} (Sonnet~4) & 0.500 & 0.045 & 0.086 & [0.500, 0.500] \\
\bottomrule
\end{tabular}%
}
\end{table}

\begin{table}[h]
\centering
\small
\setlength{\tabcolsep}{3pt}
\caption{Full multiclass retrofit-type results (Stage~2).}
\label{app:multifull}
\resizebox{\columnwidth}{!}{%
\begin{tabular}{lcccc}
\toprule
Model & Acc & F1$_w$ & F1$_m$ & AUC$_{ovr}$ \\
\midrule
AutoGluon (Ens.\ L3) & \textbf{0.702} & \textbf{0.654} & 0.348 & \textbf{0.934} \\
\LLMStacked{} & 0.680 & 0.626 & 0.281 & 0.888 \\
Random Forest (SMOTE) & 0.611 & 0.621 & \textbf{0.390} & 0.917 \\
XGBoost (SMOTE) & 0.601 & 0.614 & 0.380 & 0.914 \\
LightGBM (SMOTE) & 0.582 & 0.605 & 0.387 & 0.917 \\
CatBoost (SMOTE) & 0.589 & 0.600 & 0.356 & 0.906 \\
\LLMEmbed{} LogReg & 0.480 & 0.521 & 0.327 & 0.869 \\
Logistic Regression (SMOTE) & 0.293 & 0.332 & 0.148 & 0.768 \\
\LLMPrompted{} (Sonnet~4) & 0.100 & 0.018 & 0.012 & 0.500 \\
\bottomrule
\end{tabular}%
}
\end{table}
\newpage
\section{Monthly benchmark details}
\label{app:timeseriesdetails}
Table~\ref{app:timeseriesfull} reports the full monthly benchmark, including statistical baselines and the complete lag-feature comparison.

\begin{table}[h]
\centering
\small
\setlength{\tabcolsep}{3pt}
\caption{Aggregated monthly time-series benchmark for retrofit duration. 76 observations (60 train, 16 test).}
\label{app:timeseriesfull}
\resizebox{\columnwidth}{!}{%
\begin{tabular}{llrrr}
\toprule
Family & Model & MAE & RMSE & $R^2$ \\
\midrule
\multicolumn{5}{l}{\emph{Statistical baselines}} \\
& SMA(3) & 4.16 & 4.87 & 0.002 \\
& SMA(6) & 4.18 & 4.87 & 0.004 \\
& SES & 4.20 & 4.88 & -0.001 \\
& SBA & 4.20 & 4.88 & -0.002 \\
& ARIMA(1,1,1) & 4.19 & 4.91 & -0.014 \\
& Croston & 4.22 & 4.93 & -0.019 \\
& ETS (additive) & 4.24 & 5.82 & -0.424 \\
\midrule
\multicolumn{5}{l}{\emph{ML on lag features}} \\
& LightGBM & \textbf{3.16} & \textbf{3.70} & -1.90 \\
& AutoGluon & 3.16 & 3.79 & -1.88 \\
& SGD & 3.17 & 3.74 & -1.74 \\
& Ridge & 3.22 & 3.80 & -1.78 \\
& XGBoost & 3.33 & 4.00 & -2.14 \\
& MLP & 3.39 & 3.90 & -2.11 \\
& Random Forest & 3.44 & 3.99 & -2.30 \\
& LinearReg & 3.48 & 4.11 & -2.20 \\
\midrule
\multicolumn{5}{l}{\emph{Foundation models}} \\
& Chronos-small & 4.03 & 5.04 & -0.065 \\
& Chronos-tiny & 4.14 & 5.10 & -0.091 \\
& Chronos-base & 4.25 & 5.32 & -0.187 \\
& TIME-LLM (frozen GPT-2) & 4.62 & 5.56 & -0.217 \\
\midrule
\multicolumn{5}{l}{\emph{Ensemble}} \\
& LightGBM + AutoGluon + SGD & 3.16 & --- & -1.84 \\
\bottomrule
\end{tabular}%
}
\end{table}

\end{document}